\pdfoutput=1

\documentclass[11pt]{article}

\usepackage[final]{acl}

\usepackage{booktabs, multirow} 

\usepackage{times}
\usepackage{latexsym}

\usepackage[T1]{fontenc}

\usepackage[utf8]{inputenc}

\usepackage{microtype}

\usepackage{inconsolata}

\usepackage{graphicx}

\newcommand{\hide}[1]{}

\newcommand{\YHAMZA}{{\^{y}}}
\newcommand{\TAMARBUTA}{{$\hbar$}}

\newcommand{\DHA}{{\dh}}

\newcommand{\AYN}{{$\varsigma$}}

\newcommand{\SHADDA}{{$\sim$}}

\usepackage{arabtex}
\usepackage{utf8}
\usepackage{epstopdf}
\usepackage[T1]{fontenc}
\usepackage{hyperref}
\usepackage{xstring}

\title{Strategies for Arabic Readability Modeling}

\author{Juan Piñeros Liberato, Bashar Alhafni, Muhamed Al Khalil, Nizar Habash\\
  Computational Approaches to Modeling Language Lab\\
  New York University Abu Dhabi\\
  \texttt{\{juanpl,alhafni,muhamed.alkhalil,nizar.habash\}@nyu.edu}\\
  }
         
\begin{document}
\maketitle

\setcode{utf8}
\vocalize

\begin{abstract}
Automatic readability assessment is relevant to building NLP applications for education, content analysis, and accessibility. However, Arabic readability assessment is a challenging task due to Arabic's morphological richness and limited readability resources. In this paper, we present a set of experimental results on Arabic readability assessment using a diverse range of approaches, from rule-based methods to Arabic pretrained language models.
We report our results on a newly created corpus at different textual granularity levels (words and sentence fragments). Our results show that combining different techniques yields the best results, achieving an overall macro F\textsubscript{1} score of 86.7 at the word level and 87.9 at the fragment level on a blind test set. We make our code, data, and
pretrained models publicly available.\footnote{\url{https://github.com/CAMeL-Lab/samer-arabic-readability}}
\end{list}
\end{abstract}

\section{Introduction}
The task of automatic readability assessment aims at modeling the reading and comprehension difficulty of a given piece of text for a particular target audience. This is relevant to building and enhancing pedagogical natural language processing (NLP) applications, which aid students in language learning~\cite{xia-etal-2016-text,vajjala-meurers-2012-improving}, help teachers with designing curricula and writing assessments \cite{collins-thompson:2004}, and enable the personalization of NLP systems' output to target users with different readability levels \cite{marchisio-etal-2019-controlling,agrawal-carpuat-2019-controlling}. Research on English automatic readability assessment have garnered substantial interest in terms of dataset creation \cite{heilman-etal-2007-combining,vajjala-meurers-2013-applicability,xia-etal-2016-text,vajjala-lucic-2018-onestopenglish} and modeling advancements \cite{deutsch-etal-2020-linguistic,martinc-etal-2021-supervised,lee-vajjala-2022-neural}. In contrast, other languages such as Arabic have not received as much attention.

Arabic is a morphologically rich and orthographically ambiguous language. Words have many inflected forms varying in terms of gender, number, person, case, aspect,
mood, voice, as well as a large number of attachable clitics, such as pronominal objects and prepositions \cite{Habash:2010:introduction}. Arabic's high level of complexity poses a significant challenge for new learners. Furthermore, while Modern Standard Arabic (MSA) is used in education and the media, modern-day Arabs natively speak a variety of Arabic dialects that differ from MSA, making MSA readability a relevant issue for them too.
There are growing research efforts on Arabic readability assessment 
 \cite{Al-Khalifa:2010:automatic,AlTamimi:2014:aari,El-Haj:2016:osman,saddiki_feature_2018}. 
However, we are not aware of any work that systematically explores modeling approaches for Arabic readability at different textual granularity levels. In this paper, we present Arabic readability assessment results using diverse approaches relying on frequency and rule-based models as well as pretrained language models (PLMs). We use the newly created SAMER Arabic Text Simplification Corpus \cite{alhafni-etal-2024-samer-arabic} and report on word-level and fragment-level readability. Our contributions are as follows:

\begin{itemize}
    \item We systematically explore different modeling approaches to report on the task of Arabic readability assessment, ranging from rule-based methods to Arabic PLMs.
    \item We benchmark our models on a new corpus with different readability levels.
    \item We show that combining different modeling techniques yields optimal results: 
    86.7 word-level macro F\textsubscript{1} and 87.9 fragment-level macro F\textsubscript{1} on a blind test set.
\end{itemize}

We discuss related work in \S\ref{pwork}, provide an overview of our dataset in \S\ref{data}, describe our models for Arabic readability assessment in \S\ref{methods}, and discuss results in \S\ref{results}.

\section{Related Work}
\label{pwork}

\subsection{Readability Assessment Datasets}

Automatic readability assessment has received considerable attention, leading to the development of many resources \cite{collins-thompson-callan-2004-language,pitler-nenkova-2008-revisiting,feng-etal-2010-comparison,vajjala-meurers-2012-improving,xu-etal-2015-problems,xia-etal-2016-text,nadeem-ostendorf-2018-estimating,vajjala-lucic-2018-onestopenglish,deutsch-etal-2020-linguistic,lee-etal-2021-pushing}. Most of the English datasets were initially derived from textbooks as they are considered to be naturally suited for readability assessment research, given that the linguistic characteristics of texts become more complex as school grade increases \cite{vajjala-2022-trends}. However, many textbooks are under copyright restrictions and may not be accessible in a digitized form. This led to relying on crowd sourcing to annotate data collected from the web \cite{vajjala-meurers-2012-improving,vajjala-lucic-2018-onestopenglish} or from English assessment exams targeting second-language (L2) learners \cite{xia-etal-2016-text}, where the Common European Framework of Reference
(CEFR) \cite{cefr2001} is used.

When it comes to Arabic, specifically Modern Standard Arabic (MSA), early work on readability assessment relied mainly on academic curricula \cite{Al-Khalifa:2010:automatic,AlTamimi:2014:aari,Forsyth:2014:automatic,Khalil:2018:leveled}. More recently, there have been more efforts to create Arabic readability assessment resources. \newcite{khallaf-sharoff-2021-automatic} consolidated multiple annotated L2 datasets and mapped their readability levels to CEFR. \newcite{habash-palfreyman-2022-zaebuc} created the ZAEBUC dataset that contains essays written by native Arabic speakers, which were manually corrected and annotated for writing proficiency using the CEFR levels. \newcite{naous2023readme} introduced a manually annotated multi-domain multilingual dataset for readability assessment. In our work, we use the newly introduced publicly available SAMER Arabic Text Simplification Corpus \cite{alhafni-etal-2024-samer-arabic}, which was manually annotated for readability leveling. We discuss the corpus in more detail in \S\ref{data}. It is noteworthy that this corpus is one of the publicly available resources created by the Simplification of Arabic Masterpieces for Extensive Reading (SAMER) project which includes a readability leveled lexicon \cite{al_khalil_large-scale_2020,jiang-etal-2020-online}, and a Google Doc add-on \cite{hazim-etal-2022-arabic}.

\subsection{Approaches to Readability Assessment}
Early approaches for automatic readability assessment relied on surface-level features that could be extracted from raw text such as the average number words per sentence and the average number of characters per word. Such approaches include commonly used readability measures such as the Dale-Chall Readability Score \cite{dale:1948} and the Flesch-Kincaid Grade Level (FKGL) \cite{Flesch1948}. With the emergence of machine learning and data driven methods, approaches were extended to leverage statisical language models \cite{luo:2001} and linguistic features \cite{heilman-etal-2007-combining,Petersen:2009,ambati-etal-2016-assessing}. More recently, deep learning approaches were explored \cite{cha:2017,jiang-etal-2018-enriching,azpiazu-pera-2019-multiattentive}, including the use of Transfomer-based PLMs \cite{deutsch-etal-2020-linguistic,lee-vajjala-2022-neural,naous2023readme,imperial-kochmar-2023-automatic}.

Although a lot of this research evolved on English, approaches to modeling Arabic readability assessment witnessed a similar trend. Inspired by English readability formulas, \newcite{al-tamimi:2014} developed the Arabic Automatic Readability Index (AARI). Similarly, \newcite{El-Haj:2016:osman} introduced OSMAN, an adaptation of conventional readability formulas such as FKGL to Arabic. When it comes to machine learning models, the majority were based on linguistic features such as
perplexity scores from statistical language models \cite{Al-Khalifa:2010:automatic}, morphological information (e.g., lemmas, morphemes, part-of-speech tags) \cite{Cavalli-Sforza:2014:matching,Forsyth:2014:automatic,Saddiki:2015:text,Nassiri:2017:modern}, and syntactic features \cite{saddiki_feature_2018}. Despite the various efforts on modeling Arabic readability assessment, only few attempts were made to explore deep learning approaches. \newcite{khallaf-sharoff-2021-automatic} and \newcite{naous2023readme} presented results on using BERT \cite{devlin-etal-2019-bert,antoun-etal-2020-arabert} for readability assessment. Moreover, it is worth noting that the majority of research on Arabic readability assessment report results at either the document or sentence levels.

In our work, we draw inspiration from previous efforts to explore various modeling approaches for Arabic readability assessment at both the word and fragment levels, encompassing a spectrum from rule-based models to PLMs and their combinations.

\begin{figure*}[t!]
    \centering
    \includegraphics[width = 1.3\columnwidth]{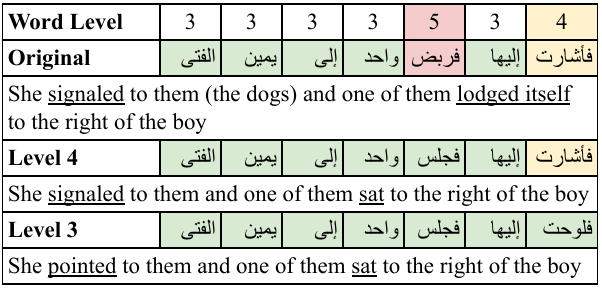}
    \captionof{figure}{An example illustrating the word-level labeling process. A word in the original text is labeled at the lowest level where it appears unchanged across the parallel versions of the text in the SAMER Corpus.}
    \label{tab:alignment}
\end{figure*}

\begin{table}[t!]
    \centering
    \begin{tabular}{lrrrrr}\toprule
&\textbf{Level 3} &\textbf{Level 4} &\textbf{Level 5} &\textbf{All} \\\midrule
\textbf{Train} &5,947 &4,543 &3,766 &14,256 \\
\textbf{Dev} &1,256 &926 &766 &2,948 \\
\textbf{Test} &1,477 &901 &776 &3,154 \\
\textbf{Total} & \textbf{8,680} &\textbf{6,370} &\textbf{5,308} &\textbf{20,358} \\
\bottomrule
\end{tabular}
    
    \captionof{table}{SAMER Corpus fragment readability level statistics per split.}
    \label{tab:fraglevels}
\end{table}

\section{Data}
\label{data}
\subsection{SAMER Corpus}
We extensively use the SAMER Arabic Text Simplification Corpus \cite{alhafni-etal-2024-samer-arabic}. The corpus consists of \textit{original} texts selected from 15 publicly available Arabic fiction novels. It includes
two simplified parallel versions for each text targeting learners at two readability levels (\textit{Level 4 }and \textit{Level 3}). 
The levels are based on \newcite{al_khalil_large-scale_2020}'s five-level lexical readability scale which ranges from Level 1 (Low Difficulty/Easy Readability) to Level 5 (High Difficulty/Hard Readability).
The SAMER Corpus simplification guidelines consider the readability level of a \textit{text} to be equal to the highest readability level found among the \textit{words} in the \textit{text}. So, a Level 4 \textit{text} cannot have any Level~5 \textit{words}, but must have at least one Level~4 word.  
As part of the manual simplification process, the human annotators simplified the original text to Level 4, and then to Level 3. This was done by first automatically obtaining the word-level readability of the original text using the SAMER Google Doc add-on \cite{hazim-etal-2022-arabic} and then manually performing minimal replacements, insertions, and deletions to simplify the text from a higher to lower readability levels. 
In some cases the annotators minimally modified some words to maintain grammatical agreement without changing their lexical readability levels.  
The add-on was used to confirm the target levels were reached; however, the annotators were allowed to overwrite incorrect automatically assigned word readability levels.
The SAMER Corpus release includes the original paragraphs and their simplified counterparts segmented into smaller parallel sentence fragments using punctuation marks. Our readability assessment experiments only use the original text fragments. The release also includes the final word readability levels; however, we do not use them as we opted to employ a more generic solution for word-level readability assignment, which we discuss next.

\begin{table}[t!]
    \centering
\begin{tabular}{lrrrrr}\toprule
\textbf{} &\multicolumn{2}{c}{\textbf{All Tokens}} &\multicolumn{2}{c}{\textbf{Train Tokens}} \\\midrule
\textbf{Level 3} &136,805 &86.6\% &97,616 &86.5\% \\
\textbf{Level 4} &14,145 &8.9\% &10,151 &9.0\% \\
\textbf{Level 5} &7,104 &4.5\% &5,056 &4.5\% \\
\textbf{Total } &\textbf{158,054} &\textbf{100\%} &\textbf{112,823} &\textbf{100\%} \\
\bottomrule
\end{tabular}
    \captionof{table}{SAMER Corpus word token readability level statistics after the word-level labeling process.}
    \label{tab:tokens}
\end{table}

\subsection{Fragments, Words, and Readability Levels}
\label{frag-word-level}
We model readability assessment on three levels: Levels 5, 4, and 3, at the word level and fragment level.
To assign a readability label to each word in the original fragments, we first obtain word-level alignments between the original fragments and the simplified parallels  using an edit distance word alignment tool \cite{alhafni-etal-2023-advancements,Khalifa:2021:Character}. We then derive the readability labels based on whether the words in the original fragments were changed in the simplified Levels 4 and 3 texts. 
See example in Figure~\ref{tab:alignment}. Similar to the SAMER Corpus, we consider the readability level of the fragment to be equal to the highest word readability level found among the words in the fragment. Tables \ref{tab:fraglevels} and \ref{tab:tokens} present the statistics of the corpus at the fragment and word levels, respectively. 

While this alignment-based approach is applicable to any parallel original-simplified text, it struggles to distinguish between lexical readability changes and grammatical agreement changes. Nevertheless, this holistic approach is valuable for text simplification tasks that require altering both words and their grammatical dependents. Importantly, this limitation does not affect the readability level of the text fragments.

\section{Approach}
\label{methods}

We present below the set of models we use for word-level and fragment-level readability labeling.

\subsection{Word-Level Readability Labeling}
\label{sec:word-level-ragg}
We investigate four models to label words according to their readability levels: Maximum Likelihood Estimation (MLE), Lexicon lookup, Frequency-based labeling, and BERT-based token classification. Each model relies on different resources as summarized in 
Table~\ref{tab:approaches}. Moreover, each model has a different set of parameters, which were tuned to optimize the performance on the Dev set.
We further investigate combining the models in a cascaded setup, leveraging their complementary strengths to address the limitations of each model individually.

\begin{figure}[t!]
    \centering
    \setlength{\tabcolsep}{1pt}
    \begin{tabular}{lcccc}\toprule
&\multicolumn{4}{c}{\textbf{Resources Used}} \\\cmidrule{2-5}
     &\textbf{SAMER}  &\textbf{CAMeL} &\textbf{CAMeL}  &\textbf{SAMER}  \\ 
\textbf{Model} &\textbf{Corpus}& \textbf{BERT} & \textbf{Tools} & \textbf{Lexicon }\\\midrule

Lexicon & &X &X &X \\
BERT &X &X & & \\
Frequency &X &Counts & & \\
MLE &X & & & \\
Default L3 & & & & \\
\bottomrule
\end{tabular}
    
    \captionof{table}{Resources used in the word-level models.}
    \label{tab:approaches}
\end{figure}

\subsubsection{Maximum Likelihood Estimation} 

The \textbf{MLE} model assigns the readability level $R$ that maximizes the  conditional probability $P(R | W)$, where $R$ is the readability level of word $W$ as estimated over the training data. 
Among in-vocabulary words, 97.2\% appear with one readability level in the training, and 0.2\% appear with all three.
For out-of-vocabulary (OOV) words, we back-off to a default readability level or to one of other models we discuss below.

\subsubsection{Lexicon} 
Our second model, \textbf{Lex},  leverages the SAMER Readability Lexicon \citep{al_khalil_large-scale_2020}, which  consists of over 40K lemmas manually annotated with their readability levels (1 to 5). For the purposes of our task, we consider lemmas of Levels 1 and 2 to be included under Level 3. During inference, we use \newcite{inoue2022morphosyntactic}'s morphological disambiguator  as implemented in CAMel Tools \cite{obeid_camel_2020} to identify the lemma and part-of-speech tag for each word. We infer the readability level of the word using its lemma's readability level in the SAMER lexicon.
In cases where the morphological disambiguator returns multiple top lemma analyses, we select the lowest readability associated with these lemmas.
For OOV words, we back-off to a default readability level or to one of other models we discuss below.

\subsubsection{Frequency-Based Models}
%
Given the limited vocabulary seen the SAMER Corpus, 
we explore different approaches to derive readability levels from frequency data building on the known observation about the inverse correlation between frequency and readability levels \cite{al-khalil-etal-2020-large}: more frequent words have easier/lower readability levels. 
We leverage our SAMER Corpus training data and link it with type frequency data  from a corpus of 12.6B word tokens (11.4M types) used to pretrain the CAMeLBERT models \citep{inoue-etal-2021-interplay}.\footnote{\url{https://github.com/CAMeL-Lab/Camel_Arabic_Frequency_Lists}} 
We sort the 11.4M word types by frequency and divide them into adjacent bins for which we assign readability levels using one of two methods:

\paragraph{Distribution-based Labeling (Dist-Freq)} 

In this method, we divide the word types into three bins that mirror the distribution of readability levels in our training data (as seen in Table~\ref{tab:tokens}). More concretely, the most frequent words that account for 86.5\% of the total distribution mass are assigned to Level 3, followed by 9.0\% assigned to Level 4, and the remaining tail assigned to Level 5. 

\paragraph{Example-based Labeling (Ex-Freq)}
Based on the assumption that types of a certain readability level tend to have similar frequencies in a large corpus, we divide the frequency-sorted types into equally sized bins based on cumulative frequency. Utilizing our training data as training examples, we assign a readability level to all the types within each bin according to the majority readability level of training words found in that bin. During inference, if words are not observed in any bin, we default to assigning a readability level of 5, reflecting the expectation that rare words are typically harder to read. We empirically experiment with different numbers of bins and found that 10,000 bins yield the highest performance in terms of macro F\textsubscript{1} score.

\subsubsection{BERT Token Classification}

We build a word-level classifier by leveraging a Transformer-based PLM. There are many Arabic monolingual
BERT \cite{devlin-etal-2019-bert} models available such as AraBERT \cite{antoun-etal-2020-arabert}, ARBERT \cite{abdul-mageed-etal-2021-arbert}, and JABER \cite{ghaddar-etal-2022-revisiting}. However, we chose to use CAMeLBERT MSA \cite{inoue-etal-2021-interplay} as it was pretrained on the largest MSA dataset to date, and following the recommendations of \newcite{inoue-etal-2021-interplay} to use it for tasks on MSA.
We fine-tune CAMeLBERT MSA using Hugging Face’s Transformers \cite{wolf-etal-2020-transformers} by adding a fully-connected linear layer with a softmax on top of its architecture. Given that BERT operates at the subword-level (i.e., wordpieces), we assign to each subword the readability level of the word it belongs to. During inference, we label each word according to the highest readability level among its subwords. We fine-tune our model on a single GPU for 10 epochs with a learning rate of 1e-5, a batch size of 32, and a maximum sequence length of 30.

\subsubsection{System Combination}
In addition to evaluating the various models discussed above, we consider their combinations to exploit their complementarities.
Our approach to combining the systems runs the \textbf{Lex} and \textbf{MLE} models first (independently and together in different orders -- four combinations) followed by one of the following \textit{six} models: default level 3, 4, or 5, \textbf{Dist-Freq}, \textbf{Ex-Freq}, or \textbf{BERT}.  The total is 24 combinations layered in two or three steps. 
See Table~\ref{tab:all-combos} in Appendix~\ref{app:layers}.
Since the early layers, Lex and MLE, do not handle unknown words, the later layers resolve these cases. 
We evaluate all these system combinations on both word and fragment leveling in terms of accuracy and macro F\textsubscript{1} score.

\subsection{Fragment-Level Readability Labeling}
We consider two approaches to fragment-level labeling: a direct BERT-based approach and an aggregation of the word-level predictions.

\subsubsection{BERT Fragment Classification}
We train a fragment-level classifier by fine-tuning CAMeLBERT MSA. We add a fully connected linear layer on top of the representation of the whole fragment. We experiment with different ways of obtaining the fragment representation from the BERT model: using the \texttt{[CLS]} token,  mean-pooling, and max-pooling, and found mean-pooling to outperform the other representations in all evaluation metrics.
We fine-tune our model using Hugging Face's Transformers \cite{Wolf:2019:huggingfaces} on a single GPU for 10 epochs with a learning rate of 5e-5, a batch size of 32, and a maximum sequence length of 20.

\subsubsection{Aggregating Word-Level Predictions}
Finally, we aggregate the word-level labels produced by the various models discussed in \S\ref{sec:word-level-ragg} above to assign fragment-level labels: the fragment label equals the highest readability level found among its words.

\begin{table*}[t!]
\centering 
    \begin{tabular}{lrrrrrrr}\toprule

        &   \textbf{Model} & \textbf{F\textsubscript{1}(3)} & \textbf{F\textsubscript{1}(4)} & \textbf{F\textsubscript{1}(5)} &      \multicolumn{1}{c}{\textbf{F\textsubscript{1}}} & \textbf{Acc.} \\\midrule

        \multirow{4}{*}{\textbf{(a)}} & Default Level 3 & 92.8 & 0.0 & 0.0  & 30.9 & 86.5 \\
                                     . & Dist-Freq & 84.2 &20.8 &28.6  & 44.5 & 71.1\\
                                      & Ex-Freq & 93.0 & 21.5 & 14.7  & 43.1 & 86.4 \\
                                      & \underline{BERT} & \underline{96.5} & \underline{67.9} & \underline{59.3} & \underline{74.6} & \underline{92.4} \\\midrule

        \multirow{4}{*}{\textbf{(b)}} & MLE → Level 3 & 95.1 &57.5 &41.2  & 64.6 & 91.0 \\
                                      & MLE → Dist-Freq &91.6 &51.0 &39.8  &60.8 &83.1 \\
                                      & MLE → Ex-Freq & 95.0 &56.9 &42.4  &64.7 &90.3 \\
                                      
                                      & \underline{MLE → BERT} &\underline{96.7} &\underline{69.9} &\underline{61.0}  & \underline{75.9} &\underline{92.8} \\\midrule

        \multirow{4}{*}{\textbf{(c)}} &Lex → Level 3 &97.8 &85.2 &74.1  &85.7 &95.7\\
                                      &Lex → Dist-Freq &97.8 &84.5 &75.2  &85.8 &95.5\\
                                      &Lex → Ex-Freq &97.8 &85.1 &74.6  &85.8 &95.7\\
                                      
                                      &\underline{Lex → BERT} &\underline{98.0} &\underline{85.1} &\underline{76.5}  &\underline{86.5} &\underline{95.9}\\\midrule

        \multirow{4}{*}{\textbf{(d)}} & Lex → MLE → Level 3 &97.8 &85.2 &74.5 &85.8  &95.8\\
                                      &Lex → MLE → BERT &98.0 &85.1 &76.5  &86.5 &95.9\\
                                      &MLE → Lex → Level 3 &98.0 &85.7 &76.9  &86.8 &96.0\\
                                      &MLE → Lex → BERT &98.1 &85.5 &78.8 &87.5 &96.2 \\
                                     &\textbf{Tuned-MLE → Lex → BERT} &\textbf{98.2} &\textbf{86.1} &\textbf{79.4} &\textbf{87.9} &\textbf{96.3}\\\bottomrule
    \end{tabular}

    \captionof{table}{Word-level results on the Dev set. F\textsubscript{1}(3), F\textsubscript{1}(4), and F\textsubscript{1}(5) are the macro F\textsubscript{1} scores for levels 3, 4, and 5, respectively. F\textsubscript{1} is the overall macro F\textsubscript{1} score. Underlined numbers represent the best results in each subcategory of experiments. Best overall results are in bold. }
    \label{tab:wordwise}

\end{table*}

\begin{table*}[t!]
    \centering
    \begin{tabular}{lrrrrrr}\toprule
            \textbf{Model} & \textbf{F\textsubscript{1}(3)} &\textbf{F\textsubscript{1}(4)} &\textbf{F\textsubscript{1}(5)}  &\multicolumn{1}{c}{\textbf{F\textsubscript{1}}} &\textbf{Acc.} \\\midrule

            BERT (Fragment-Level) &88.7 &68.2 &58.8  &71.9 &79.4\\
            BERT (Word-Level) &85.1 &70.2 &69.4  &74.9 &76.4\\
            Lex → MLE → Level 3 &90.9 &85.0 &81.1  &85.7 &86.6\\
            Lex → MLE → BERT &91.2 &85.3 &82.9  &86.4 &87.2\\
            MLE → Lex → Level 3 &91.2 &85.9 &83.0  &86.7 &87.5\\
            MLE → Lex → BERT &91.6 &86.1 &84.6 &87.5 &88.1 \\
            \textbf{Tuned-MLE → Lex → BERT} &\textbf{92.1} &\textbf{86.7} &\textbf{84.9} &\textbf{87.9}  &\textbf{88.6}\\\bottomrule
    \end{tabular}
    \captionof{table}{Fragment-level results on the Dev set. }
    \label{tab:fragwise}
\end{table*}

\section{Results}
\label{results}
We present and discuss the results of our evaluation below. The complete set of results for word-level and fragment-level labeling across all experimental setups is available in Appendix~\ref{app:layers}.

\subsection{Word-Level Labeling Results}
Table~\ref{tab:wordwise} presents the results on the Dev set. We start off with the results of the standalone models in Table~\ref{tab:wordwise}\textbf{(a)}. The frequency-based approaches (Dist-Freq and Ex-Freq) improve over the majority class baseline (Default Level 3). However, they are outperformed by BERT. This improvement is attributed to the significant increase in the F\textsubscript{1} scores for Level 4 and Level 5 words. 
In Table~\ref{tab:wordwise}\textbf{(b)} we show that the results improve further when combining the MLE model with BERT as a back-off system. 

Results in Table~\ref{tab:wordwise}(\textbf{c}) show that using the frequency-based and BERT models as back-off systems to Lex improve the results compared to defaulting to Level 3, with  Lex → BERT being the best performer. However,
the improvements when using a back-off model to Lex are not as large as the ones observed when using the MLE model (Table~\ref{tab:wordwise}\textbf{(b)}). This is due to the larger coverage the Lexicon has on the Dev set  (96.4\% of all tokens) compared to the MLE system (79.0\%).

Finally, in Table~\ref{tab:wordwise}\textbf{(d)} we present the maximal combination results. We find that using the MLE model, followed by Lex and then BERT yields the best results. 
We further tune this combination by considering different probability thresholds at which to back-off from MLE. We found 85\% MLE minimum probability to give the best results on the Dev set. 
Our best model combination is thus
\textbf{Tuned-MLE → Lex → BERT} with 87.9 F\textsubscript{1}.

\begin{table}[t!]
    \centering
    %

    \setlength{\tabcolsep}{6pt}
    \begin{tabular}{ccrrrr}\toprule
\textbf{Fragment} &\textbf{Word} & \multicolumn{3}{c}{\multirow{2}{*}{\textbf{\# of Fragments}}} \\
\textbf{Label} &\textbf{Errors} & & \\\midrule
Correct &0 & 2,300 &78.\% &78.0\% \\\midrule
Correct &1 &239 &8.1\% &\multirow{3}{*}{10.6\%} \\
Correct &2 &56 &1.9\% & \\
Correct &3+ &16 &0.5\%& \\\midrule
Incorrect &1 &283 &9.6\% &\multirow{3}{*}{11.4\%} \\
Incorrect &2 &38 &1.3\% & \\
Incorrect &3+ &16 &0.5\% & \\
\bottomrule
\end{tabular}
    \captionof{table}{Summary of fragment-word error combinations on the Dev set.
    We identify  three groups: correct fragments with no word errors, correct fragments with some word errors, and incorrect fragments with word errors.
    }
    \label{tab:errorsummary}
\end{table}

\subsection{Fragment-Level Labeling Results}

Table~\ref{tab:fragwise} presents the fragment-level results on the Dev set.
We find that, although the fragment-level BERT classifier does better than its word-level counterpart, the aggregated word-level models perform better on the fragment-level. We obtain the best results using the (\textbf{Tuned-MLE → Lex → BERT}) model, achieving an F\textsubscript{1} score of 87.9. It is interesting to note that the best system coincidentally achieves the same overall F\textsubscript{1} macro at the word and fragment levels. 
Our best system is better at predicting Level~3 words compared to Level~3 fragments (98.2 v.s. 92.1). Conversely, the system is better at predicting Level~4 and Level~5 fragments compared to the words. This makes sense given that Level~3 fragments are exclusively composed of Level~3 words, any word-level error on a Level~3 fragment leads to fragment error. 
In sum, the 3.7\% accuracy errors at the word level lead to 11.4\% accuracy errors at the fragment level.
Table~\ref{tab:errorsummary} presents a detailed breakdown of the combinations of word and fragment errors.


Finally, we revisit our best model combination 
\textbf{Tuned-MLE → Lex → BERT} in 
Table~\ref{tab:decomposition}, where we give a summary of the decisions and mistakes made by each of its three components and their effect on word-level and fragment-level performance.
We notice that most of the decisions were taken by the MLE model, which had the lowest error rate, and the lowest rate of error propagation to the fragment level. However, when errors at the word level happen, there is a large chance a fragment error will follow suit in all three models. Moreover, we note that the performance is highly degraded by the last model (BERT) decisions, with 35.6\% word-level and 67.7\% fragment-level errors.

\subsection{Blind Test Results} 
Table \ref{tab:testresults} presents the results on the Test set. We observe consistent conclusions to the Dev results. Our best system (\textbf{Tuned-MLE → Lex → BERT}) achieves an overall F\textsubscript{1} score of 86.7 at the word level and 87.9 at the fragment level.

\begin{table}[t!]
    \centering
    %
    \setlength{\tabcolsep}{5pt}
    \begin{tabular}{lrrrr}\toprule
\textbf{} &\textbf{MLE} &\textbf{LEX} &\textbf{BERT} \\\midrule
\textbf{Decisions} &17,058 &4,843 &174 \\
\textbf{Mistakes} &377 &375 &62 \\\hline
\textbf{Applied } &77.3\% &21.9\% &0.8\% \\
\textbf{Word Error } &2.2\% &7.7\% &35.6\% \\
\textbf{Fragment Error } &41.4\% &56.3\% &67.7\% \\
\bottomrule
\end{tabular}
    
    \captionof{table}{The word-level decisions taken by each of the layers of the best-performing system on the Dev set's 22,075 tokens, and their error rates in terms of word-level and fragment-level labeling.}
    \label{tab:decomposition}
\end{table}

\subsection{Manual Error Analysis}

We manually classified 100 cases of word readability errors from the Dev set (out of 814 or 3.7\% of all words) into seven distinct error types. We provide a brief description of each error type below, with its percentage of occurrence. The errors are presented in order of precedence, so if there is an Input error, we do not consider any other error below it, and so on.

\begin{table*}[t!]
    \centering
    \setlength{\tabcolsep}{4pt}
    \begin{tabular}{lrrrrr|rrrrrr}\toprule
    \multirow{2}{*}[-3pt]{\textbf{Model}} &  \multicolumn{5}{c|}{\textbf{Word-Level}} &  \multicolumn{5}{c}{\textbf{Fragment-Level}} \\\cmidrule{2-11}
     &  \textbf{F\textsubscript{1}(3)} &\textbf{F\textsubscript{1}(4)} &\textbf{F\textsubscript{1}(5)}  &\multicolumn{1}{c}{\textbf{F\textsubscript{1}}} &\textbf{Acc.} &  \textbf{F\textsubscript{1}(3)} &\textbf{F\textsubscript{1}(4)} &\textbf{F\textsubscript{1}(5)}  &\multicolumn{1}{c}{\textbf{F\textsubscript{1}}} &\textbf{Acc.} \\\midrule

    BERT &  96.8 &  70.4 &  56.9  &  74.7 &  92.9 &  87.9 &  71.8 &  66.8  &  75.5 &  78.3 \\

    MLE → BERT &  96.9 &  71.9 &  58.0  &  75.6 &  93.2 &  88.7 &  73.4 &  67.5  &  76.5 &  79.3 \\

    Lexicon → BERT &  97.8 &  84.8 &  72.5  &  85.0 &  95.6 &  91.1 &  85.6 &  79.4  &  85.4 &  86.7 \\

    \textbf{Tuned-MLE → Lex → BERT} &  \textbf{98.1} &  \textbf{86.2} &  \textbf{75.9}  &  \textbf{86.7} &  \textbf{96.2} &  \textbf{93.3} &  \textbf{87.3} &  \textbf{82.9}  &  \textbf{87.9} &  \textbf{89.1} \\\bottomrule
    \end{tabular}
    \captionof{table}{Results on the Test set at both the word and fragment levels.}
    \label{tab:testresults}
\end{table*}

    \paragraph{Input Error: 3\%} The word is malformed in terms of spelling; e.g., \<معارفة> \textit{m{\AYN}Arf{\TAMARBUTA}} instead of \<معارفه> \textit{m{\AYN}Arfh} `his features'.

        \paragraph{Gold Reference Annotation Error: 18\%} The human annotator made a mistake of under-simplification or over-simplification, e.g., rewriting \<فقطعا أكثر الطريق> `they \textit{crossed} most of the road' as  \<فمشيا أكثر الطريق> `they \textit{walked} most of the road' (L4) is unnecessary since the \textit{original} is not L5.

    \paragraph{Gold Reference Determination Error: 8\%} 
    As discussed in \S\ref{frag-word-level},
    our process to determine the word-level readability confused grammatical agreement changes with lexical simplification changes, e.g., the phrase 
\<ملامحه المتجعدة>  \textit{mlAmHh Almtj{\AYN}d{\TAMARBUTA}} `his wrinkled features' is simplified correctly to 
\<وجهه المتجعد>  \textit{wjhh Almtj{\AYN}d} `his wrinkled face' by changing the first word's lemma and only changing the gender agreement of the second word; however both are considered changed and thus assigned a higher level.

     \paragraph{MLE Error: 22\%} The MLE model misclassified a word, e.g., confusing \<مهد> \textit{mahd} `cradle' (L5) with the verb \textit{mah{\SHADDA}ad} `he paved' (L4).
     
     \paragraph{Disambiguation Error: 31\%} The Lex model misclassified a word  whose lemma is in the lexicon, because of morphosyntactic or lemmatization choice errors, e.g., \<منفذ> \textit{manfa{\DHA}} `outlet' (L4) is incorrectly identified as \textit{munaf{\SHADDA}i{\DHA}} `executor' (L3).

     \paragraph{Lexicon Error: 11\%} The correct lemma is not in the lexicon, and an incorrect lemma is chosen, e.g., for the word \<لامسا> \textit{lAmsA} the system chose the verbal analysis \textit{laAmas} `touched' instead of the nominal active participle \textit{laAmis} `touching'.

     \paragraph{BERT Error: 7\%} The word is OOV in the lexicon, and BERT misclassified it, e.g., the lemma \<حائم> \textit{HA{\YHAMZA}m} `hovering' (annotators assigned L5) is not in the lexicon, and BERT misclassified it as L4.

 The lexicon and disambiguation errors take a significant share of all errors and direct us towards working on improving these resources in the future; as better generalizing models are developed, we would rely less on the MLE model.  The rate of gold errors is low and within reason given the complexity of the task.

\section{Conclusions and Future Work}
We explored the problem of Arabic readability assessment using a diverse set of approaches relying on frequency and rule-based models as well as Arabic pretrained language models (PLMs). We reported results using a newly manually created corpus at both the word and fragment levels. We further highlighted the strengths and weaknesses of each approach and underscored the importance of employing different strategies to address Arabic readability assessment effectively. Our findings demonstrate that combining different modeling techniques yields the best results, achieving an overall macro F\textsubscript{1} score of 86.7 at the word level and 87.9 at the fragment level.

In future work, we plan to explore the effect of various linguistic features in enhancing machine learning models for Arabic readability assessment. We plan to continue to improve basic enabling technologies such as morphological disambiguation and lemmatization and study their effect on readability models.
We further plan to employ our best results in the development of online tools to support pedagogical NLP applications.

\section*{Acknowledgements}
 We acknowledge the support of the High Performance Computing Center at New York University Abu Dhabi. 

\section*{Limitations}
We acknowledge the following limitations.
\begin{itemize}
    \item By focusing on lexical readability, the approach used to create the SAMER corpus ignores many readability related phenomena such as phonological, morphological and syntactic complexity.

    \item The SAMER corpus does not cover all variations of Arabic text genres, which limits the robustness of the results.
 
    \item The assessment at three readability levels might not capture the full complexity of text readability at wider age and education level ranges. 

    \item The study lacks human evaluation to corroborate the automatic readability assessments, which is crucial for validating the practical effectiveness of the models. 
\end{itemize}

\bibliography{custom,anthology,camel-bib-v3, JuanBib}
\onecolumn

\appendix

\clearpage

\section{All Word-level Model Combination  Results}
\label{app:layers}

\begin{table}[!h]
\centering 
\tabcolsep4pt
    \begin{tabular}{l|ccccc|cccccc}\toprule
    \multirow{2}{*}[-3pt]{\textbf{Model}}  & \multicolumn{5}{c|}{\textbf{Word-Level}} & \multicolumn{5}{c}{\textbf{Fragment-Level}} \\
     \cmidrule{2-11}
  & \textbf{F\textsubscript{1}(3)} & \textbf{F\textsubscript{1}(4)} & \textbf{F\textsubscript{1}(5)}  & \textbf{F\textsubscript{1}} & \textbf{Acc.} & \textbf{F\textsubscript{1}(3)} & \textbf{F\textsubscript{1}(4)} & \textbf{F\textsubscript{1}(5)}  & \textbf{F\textsubscript{1}} & \textbf{Acc.} \\
 \midrule

        Default level 3 & 92.8 & 0 & 0 & 30.9 & 86.5 & 59.8 & 0 & 0 & 19.9 & 42.6 \\

        Default level 4 & 0 & 16.3 & 0 & 5.4 & 8.9 &  0 & 47.8 & 0 & 15.9 & 31.4 \\

        Default level 5 & 0 & 0 & 8.7 & 2.9 & 4.6 & 0 & 0 & 41.2 & 13.7 & 26.0 \\

        Dist-Freq  &  84.2 & 20.8 & 28.6 & 44.5 & 71.1 & 39.2 & 27.1 & 49.0  & 38.4 & 40.3 \\

        Ex-Freq & 93.0 & 21.5 & 14.7 & 43.1 & 86.4 & 64.9 & 29.3 & 29.5 & 41.3 & 50.5 \\

        BERT  & 96.5 & 67.9 & 59.3 & 74.6 & 92.4 & 85.1 & 70.2 & 69.4 & 74.9 & 76.4 \\
\hline
        MLE → L3  &  95.1 & 57.5 & 41.2 & 64.6 & 91.0 & 75.1 & 62.1 & 50.2 & 62.5 & 67.1 \\

        MLE → L4  &  89.9 & 46.0  & 41.2 & 59.0 &  81.0 & 54.6 & 57.6 & 50.2 & 54.1 & 55.4 \\

        MLE → L5  &  89.9 & 57.5 & 29.8 & 59.1 & 79.7 & 54.6 & 26.6 & 49.4 & 43.5 & 46.7 \\

        MLE → Dist-Freq & 91.6 & 51.0 & 39.8 & 60.8 & 83.1 & 60.5 & 44.0 & 55.9 & 53.5 & 54.2 \\

        MLE → Ex-Freq  &  95.0 & 56.9 & 42.4 & 64.7 & 90.3 & 75.7 & 60.8 & 56.0 & 64.2 & 67.3 \\

        MLE → BERT & 96.7 & 69.9 & 61.0 & 75.9 & 92.8 & 85.4 & 71.0 & 70.7 & 75.7 & 77.1 \\

        MLE → Lex → L3 & 98.0 & 85.7 & 76.9 & 86.8 & 96.0 & 91.2 & 85.9 & 83.0 & 86.7 & 87.5 \\

        MLE → Lex → L4 & 98.0 & 82.9 & 76.9 & 85.9 & 95.8 & 91.0 & 83.3 & 83.0 & 85.8 & 86.5 \\

        MLE → Lex → L5 & 98.0 & 85.7 & 77.5 & 87.1 & 96.0 & 91.0 & 85.5 & 83.5 & 86.7 & 87.3 \\

        MLE → Lex → Dist-Freq  &  98.0 &  85.5 & 78.0 & 87.2 & 96.0 & 91.4 & 85.9 & 84.1 & 87.2 & 87.8 \\

        MLE → Lex → Ex-Freq & 98.0 & 85.6 & 77.2 & 86.9 & 96.0 & 91.3 & 85.8 & 83.4 & 86.9 & 87.6 \\

        MLE → Lex → BERT  &  98.1 & 85.5 & 78.8 & 87.5 & 96.2 & 91.6 & 86.1 & 84.6 & 87.4 & 88.1 \\

        Lex → L3 & 97.8 & 85.2 & 74.1 & 85.7 & 95.7 & 90.7 & 85.0 & 80.9 & 85.5 & 86.5 \\

        Lex → L4 & 97.3 & 78.6 & 74.1 & 83.4 & 94.6 & 88.6 & 80.3 & 80.9 & 83.2 & 83.9 \\

        Lex → L5 & 97.3 & 85.2 & 68.8 & 83.8 & 94.9 & 88.6 & 82.3 & 77.5 & 82.8 & 83.5 \\

        Lex → Dist-Freq & 97.8 & 84.5 & 75.2 & 85.8 & 95.5 & 90.6 & 84.4 & 82.1 & 85.7 & 86.3 \\

        Lex → Ex-Freq  &  97.8 & 85.1 & 74.6 & 85.8 & 95.7 & 90.9 & 84.9 & 81.5 & 85.8 & 86.6 \\

        Lex → BERT & 98.0 & 85.1 & 76.5 & 86.5 & 95.9 & 91.3 & 85.4 & 83.0  & 86.6 & 87.3 \\

        Lex → MLE → L3 & 97.8 & 85.2 & 74.5 & 85.8 & 95.8 & 90.9 & 85.0 & 81.1 & 85.7 & 86.6 \\

        Lex → MLE → L4 & 97.9 & 82.5 & 74.5 & 84.9 & 95.5 & 90.8 & 82.4 & 81.1 & 84.8 & 85.6 \\

        Lex → MLE → L5 & 97.9 & 85.2 & 75.3 & 86.1 & 95.7 & 90.8 & 84.9 & 82.3 & 86.0 & 86.7 \\

        Lex → MLE → Dist-Freq  &  97.9 & 85.1 & 75.8 & 86.2 & 95.7 & 91.1 & 85.1 & 82.6 & 86.3 & 87.0 \\

        Lex → MLE → Ex-Freq & 97.8 & 85.1 & 74.8 & 85.9 & 95.7 & 91.1 & 84.9 & 81.6 & 85.9 & 86.7 \\

        Lex → MLE → BERT  &  98.0 & 85.1 & 76.5 & 86.5 & 95.9 & 91.2 & 85.3 & 82.9 & 86.4 & 87.2 \\
\hline
        \textbf{Tuned-MLE → Lex → BERT} & \textbf{98.2} & \textbf{86.1} & \textbf{79.4}  & \textbf{87.9} & \textbf{96.3} & \textbf{92.1} & \textbf{86.7} & \textbf{84.9}  & \textbf{87.9} & \textbf{88.6} \\\bottomrule

\end{tabular}
\caption{Word-level results on the Dev set for all the layered experiments.}
\label{tab:all-combos}
\end{table}

\end{document}